\documentclass[10pt,journal,compsoc,twoside]{IEEEtran}
 \usepackage{subcaption} %cris: cannot compile if it is not the first package
 \usepackage{arabtex}
 \usepackage{utf8}
 \usepackage[T1,T2A]{fontenc}
 \usepackage[utf8]{inputenc} 
 \usepackage[english]{babel}
 \usepackage{url}
 \usepackage{color}
 \usepackage{footnote}

\usepackage[normalem]{ulem}

\usepackage{enumitem}

\ifCLASSOPTIONcompsoc
  \usepackage[nocompress]{cite}
\else
  \usepackage{cite}
\fi
\ifCLASSINFOpdf
  \usepackage[pdftex]{graphicx}
  \DeclareGraphicsExtensions{.pdf,.jpeg,.png}
\else
\fi

\usepackage{amsmath}

\usepackage{booktabs}
\usepackage{multirow}
\newcommand*\rot{\rotatebox{90}}
\newcommand{\en}{$en$}
\newcommand{\es}{$es$}
\newcommand{\ar}{$ar$}
\newcommand{\fr}{$fr$}
\newcommand{\de}{$de$}

\newcommand{\sysarw}{\texttt{S1-w}}
\newcommand{\sysarl}{\texttt{S1-l}} 
\newcommand{\sysfrwdl}{\texttt{S2-w-d512}}
\newcommand{\sysfrwdm}{\texttt{S2-w-d1024}}
\newcommand{\sysfrwdh}{\texttt{S2-w-d2048}}
\newcommand{\ssysfrwdl}{\texttt{S2-w-d512}}
\newcommand{\ssysfrwdm}{\texttt{S2-w-d1024}}
\newcommand{\ssysfrwdh}{\texttt{S2-w-d2048}}

\newcommand{\weCCCnmt}{\texttt{WE-d300-nmt}}
\newcommand{\weMXXIVnmt}{\texttt{WE-d1024-nmt}}

\newcommand{\sts}{subSTS2017}
\newcommand{\news}{newstest2013}

\newcommand{\trad}{trad}
\newcommand{\sims}{semrel}
\newcommand{\unrel}{unrel}

\newcommand{\deltaTrUr}{$\Delta_{\rm tr-ur}$}

\newcommand{\Ni}{({\em i})~}
\newcommand{\Nii}{({\em ii})~}
\newcommand{\Niii}{({\em iii})~}
\newcommand{\Niv}{({\em iv})~}
\newcommand{\Nv}{({\em v})~}

\begin{document}
\title{An Empirical Analysis of NMT-Derived \\ Interlingual Embeddings and their Use in \\ Parallel Sentence Identification}


\author{Cristina~Espa\~na-Bonet,
        \'Ad\'am~Csaba~Varga,
        Alberto~Barr\'on-Cede\~no,
        and~Josef~van~Genabith% <-this % stops a space
% \author{Michael~Shell,~\IEEEmembership{Member,~IEEE,}
%         John~Doe,~\IEEEmembership{Fellow,~OSA,}
%         and~Jane~Doe,~\IEEEmembership{Life~Fellow,~IEEE}% <-this % stops a space
\IEEEcompsocitemizethanks{%
    \IEEEcompsocthanksitem Cristina Espa\~na-Bonet and Josef van~Genabith are with 
%     Deutsches Forschungszentrum f\"ur Kuenstliche Intelligenz, 
    DFKI and University of Saarland, Saarbr\"ucken, Germany\protect\\
    E-mail: cristinae@dfki.de
    \IEEEcompsocthanksitem \'Ad\'am~Csaba Varga is with University of Saarland, Saarbr\"ucken, Germany
    \IEEEcompsocthanksitem Alberto Barr\'on-Cede\~no is with Qatar Computing Research Institute, HBKU, Qatar}% <-this % stops an unwanted space
    \thanks{Manuscript received April 1, 2017; revised September 24, 2017; accepted October 1, 2017. Date of publication October 18, 2017; date of current version November 16, 2017. The guest editor coordinating the review of this paper and approving it for publication was Dr. Mary P. Harper. Corresponding~author: Cristina Espa\~na-Bonet.}
    \thanks{Digital Object Identifier 10.1109/JSTSP.2017.2764273}

}


\markboth{To appear: ~~IEEE JOURNAL OF SELECTED TOPICS IN SIGNAL PROCESSING, Vol.~11, No. 8, December~2017}%
{ESPA\~NA-BONET et al.: EMPIRICAL ANALYSIS OF NMT-DERIVED INTERLINGUAL EMBEDDINGS}

\IEEEpubid{~This work is licensed under a Creative Commons Attribution 3.0 License. For more information, see http://creativecommons.org/licenses/by/3.0/}

\setcounter{page}{1340}
\IEEEtitleabstractindextext{%
\begin{abstract}
End-to-end neural machine translation has overtaken statistical machine translation in terms of translation quality for some language pairs, specially those with large amounts of parallel data. Besides this palpable improvement, neural networks provide several new properties. A single system can be trained to translate between many languages at almost no additional cost 
other than training time. Furthermore, internal representations learned by the network serve as a new semantic representation of words ---or sentences--- which, unlike standard word embeddings, are learned in an essentially bilingual or even 
multilingual context.
In view of these properties, the contribution of the present work is two-fold. 
First, we systematically study the NMT context vectors, i.e.\ output of the encoder, and their power as an interlingua representation of a sentence. We assess their quality and effectiveness by measuring similarities across translations, as well as semantically related and semantically unrelated sentence pairs. 
Second, as extrinsic evaluation of the first point, we identify parallel sentences in comparable corpora, obtaining an $F_1=98.2\%$ on data from a shared task when using only NMT context vectors. Using context vectors jointly with similarity measures $F_1$ reaches $98.9\%$.
\end{abstract}

% % Note that keywords are not normally used for peerreview papers.
% \begin{IEEEkeywords}
% Computer Society, IEEE, IEEEtran, journal, \LaTeX, paper, template.
% \end{IEEEkeywords}
}

\maketitle

\IEEEdisplaynontitleabstractindextext

% \fi
\IEEEpeerreviewmaketitle

%% Introduction for the article
%% JSTSP Special Issue on End-to-End Speech and Language Processing

\IEEEraisesectionheading{\section{Introduction}\label{s:intro}}

\IEEEPARstart{E}{nd}-to-end neural machine translation systems (NMT) emerged in 2013~\cite{kalchbrennerBlunson:2013} as a promising alternative to statistical and rule-based systems. Nowadays, they are the state of the art for language pairs with large amounts of parallel data~\cite{WMT1:2016,WMT2:2017} and have nice properties that other paradigms lack. 
We highlight three: being a deep learning architecture, NMT does not require manually predefined features; it allows for the simultaneous training of systems across multiple languages; and it can provide zero-shot translations, i.e.\ translations for language pairs not directly seen in the training data~\cite{haEtal:2016,johnsonEtal:2016}.

Multilingual neural machine translation systems (ML-NMT) have interesting features. To perform multilingual translation, the network must project all the languages into the same common embedding space. In principle this space is multilingual, but the network does more than simply locating words according to their language and meaning independently. Previous studies suggest that the network locates words according to their semantics, irrespective of their language~\cite{sutskeverEtal:2014,haEtal:2016,johnsonEtal:2016}. That is somehow reinforced by the fact that zero-shot translation is possible (though at low quality). If that is confirmed, ML-NMT systems are learning a representation akin to an interligua for a source text and such interlingual embeddings could be used to assess cross-language similarity, among other applications. 

In the past, the analysis of internal embeddings in NMT systems has been limited to visualisations; e.g., showing the proximity between semantically-similar representations.
In the first part of this paper, we go beyond graphical analyses and search for empirical evidence of interlinguality. We address four specific research questions.
RQ1: Whether the embedding learned by the network for a source text also depends on the target language. 
RQ2: How distinguishable representations of semantically-similar and semantically-distant sentence pairs are.
RQ3: How close representations of sentence pairs within and across languages are. 
RQ4: How representations evolve throughout the training.
These questions are addressed by means of statistics on cosine similarities between pairs of sentences both in a monolingual and a cross-language setting. 
In order to do that, we perform a large number of experiments using parallel and comparable data in Arabic, English, French, German, and Spanish (\ar, \en, \fr, \de, and \es\ onwards). The second part of the paper is devoted to an application of the findings gathered in the first part: we explore the use of the ``interlingua'' representations to extract parallel sentences from comparable corpora. In this context, comparable corpora are text data on the same topic that are not direct translations of each other but may contain fragments that are translation equivalents; e.g., Wikipedia or news articles on the same subject in different languages.
We evaluate the performance of supervised classification algorithms based upon our best contextual representations when discriminating between parallel and non-parallel sentences. 

The article is organised as follows. 
Section~\ref{s:architecture} overviews the architecture of NMT systems. Section~\ref{s:related} describes the related work. Section~\ref{s:data} details the ML-NMT engines used in our analysis, presented in Section~\ref{s:vectors}.  
Section~\ref{s:adaptation} presents a use case: using the embeddings to identify parallel sentences. The conclusions are drawn in Section~\ref{s:conclusions}.

\section{Background}
\label{s:architecture}

State-of-the-art NMT systems use an encoder--decoder architecture with recurrent neural networks (RNN)~\cite{choEtal:2014,sutskeverEtal:2014,bahdanauEtal:2014}. The encoder projects source sentences into an embedding space. The decoder generates target sentences from the encoder embeddings.
Let $s=(x_1, \dots,x_n)$ be a source sentence of length $n$. The encoder encodes $s$ as a set of context vectors%
\footnote{Called ``annotation vectors'' by~\cite{bahdanauEtal:2014}, who use ``context vectors'' to designate the vectors after the attention mechanism.}, one per word: 
\begin{equation}
  \mathbf{c} = \left\{ \mathbf{h}_1, \mathbf{h}_2, \ldots, \mathbf{h}_n \right\}.
  \label{e:ctx_word}
\end{equation}
Each component of this vector is obtained by concatenating the forward ($\overrightarrow{\mathbf{h}}_i$) and backward ($\overleftarrow{ \mathbf{h}}_{i}$) encoder RNN hidden states:
\begin{align}
  \mathbf{h}_i = & \left[ \overleftarrow{\mathbf{h}}_i, \overrightarrow{\mathbf{h}}_i \right] \\
                 = & \left[ {f(\overleftarrow{\mathbf{h}}_{i-1}, \mathbf{r}_i )} , {f(\overrightarrow{\mathbf{h}}_{i+1},  \mathbf{r}_i )}    
             \right] ,
\end{align}
where $f$ is a recurrent unit (GRU: Gated Recurrent Units~\cite{choEtal:2014} in our experiments)
and $\mathbf{r}_i$ is the embedding space representation of the source word at position $i$:
$\mathbf{r}_i= \mathbf{W_x}\cdot \mathbf{x}_i$.

The decoder generates the output sentence $t=(y_1,\dots,y_m)$ of length $m$ on a word-by-word basis. The recurrent hidden state of the decoder $\mathbf{z}_j$ is computed using its previous hidden state $\mathbf{z}_{j-1}$, as well as the previous continuous representation of the target word $\mathbf{t}_{j-1}$ and the weighted context vector $\mathbf{q}_j$ at time step $j$:
\begin{align}
\mathbf{z}_j     & =  g(\mathbf{z}_{j-1},\mathbf{t}_{j-1},\mathbf{q}_j)\\
\mathbf{t}_{j-1} & =  \mathbf{W_y} \cdot \mathbf{y}_{j-1} ,
\end{align}
where $g$ is a non-linear function and $\mathbf{W_y}$ is the matrix of the target embeddings. The weighted context vector $\mathbf{q}_j$ is calculated by the \textit{attention mechanism} as described in \cite{bahdanauEtal:2014}. Its function is to assign weights to the context vectors in order to selectively focus on different source words at different time steps of the translation. To this end, a single-hidden-layer feed-forward neural network is utilised to assign relevance scores ($a$, as they can be interpreted as alignment scores) to the context vectors, which are then normalised into probabilities by the \textit{softmax} function:
\begin{align}
a(\mathbf{z}_{j-1}, \mathbf{h}_i)=\mathbf{v}_a \cdot  \mathrm{tanh}(\mathbf{W}_a \cdot  \mathbf{z}_{j-1}+\mathbf{U}_a \cdot  \mathbf{h}_i)\\
\alpha_{ij}=\mathrm{softmax\left( a(\mathbf{z}_{j-1}, \mathbf{h}_i) \right)} ,\quad 
\mathbf{q}_j=\sum\limits_i\alpha_{ij}\mathbf{h}_i
\end{align}
The attention mechanism takes the decoder's previous hidden state $\mathbf{z}_{j-1}$ and the context vector $\mathbf{h}_i$ as inputs and weighs them up with the trainable weight matrices $\mathbf{W}_a$ and $\mathbf{U}_a$, respectively.
Finally, the probability of a target word is given by the following softmax activation~\cite{nematus}:
\begin{flalign}
p(y_j|\mathbf{y}_{<j}, \mathbf{x}) = p(y_j|\mathbf{z}_{j},\mathbf{t}_{j-1},\mathbf{q}_j) =  \mathrm{softmax} \left( \mathbf{p}_j \mathbf{W}_o \right), \\
\mathbf{p}_j = \mathrm{tanh} \left( \mathbf{z}_j \mathbf{W}_{p1} + \mathbf{W_y}[y_{j-1}] \mathbf{W}_{p2} + \mathbf{q}_j \mathbf{W}_{p3} \right)
\end{flalign}
where $\mathbf{W}_{p1}, \mathbf{W}_{p2}, \mathbf{W}_{p3}, \mathbf{W}_o$ are trainable matrices.

A number of papers extend this architecture to perform multilingual translation. They use multiple encoders and/or decoders with multiple or shared attention mechanisms~\cite{luongEtal:2015,dongEtal:2015,firatEtAla:2016,zophKnight:2016,leeEtal:2016}. 
A simpler approximation~\cite{haEtal:2016,johnsonEtal:2016} considers exactly the same architecture as the one-to-one NMT for many-to-many NMT using multilingual data with some additional labelling. The authors in \cite{johnsonEtal:2016} append the tag of the target language to the source-side sentences, forcing the decoder to translate to the appropriate language. The authors in \cite{haEtal:2016} also include tags specifying the language of every source word. Both papers show how these ML-NMT architectures can improve the translation quality between under-resourced language pairs and how they can be used for zero-shot translation.
Given the premise that the encoder of an NMT system projects sentences into an embedding space, we can expect the encoder of ML-NMT systems to project sentences in different languages into a common (interlingual) embedding space. One of our aims is to study the characteristics of the internal representations of the encoder module in a ML-NMT system, and validate this assumption (see Section~\ref{s:vectors}).

\section{Related Work}
\label{s:related}

% previous qualitative analyses
There is some relevant previous research on qualitative studies of the NMT embedding space. The authors in \cite{sutskeverEtal:2014} show how a monolingual NMT encoder represents sentences with similar meaning close in the embedding space. They show graphically ---with two instance sentences--- that clustering by meaning goes beyond a bag-of-words understanding, and that differences caused by the order of the words are reflected in the representation. 
The authors in \cite{haEtal:2016} go one step further and visualise the internal space in a many-to-one language NMT system.
A 2D-representation of some multilingual word embeddings from the encoder after training displays translations and related words close together. 
Experiments in~\cite{johnsonEtal:2016} provide visual evidence of a shared space for the attention vectors in a ML-NMT setup. Sentences with the same meaning but in different languages group together, except for zero-shot translations. When a language pair has not been seen during training, the embeddings lie in a different region of the space. 
In~\cite{johnsonEtal:2016} the authors study the representation generated by the \emph{attention vectors}; i.e.\ the vectors showing the activations in the layer between encoder and decoder. The activations indicate which part of the source sentence is important during decoding to produce a particular chunk of the translation. Although the attention mechanism is shared across all the languages, the relevant chunks in the source sentence can vary depending on the target language.

% what do we do and how our work differs
In contrast to previous qualitative research, we focus on the \emph{context vectors}: the concatenation of the hidden states of the forward and the backward network in the encoding module ---right before applying the attention mechanism. Our goal goes beyond understanding the internal representations learned by the network: we aim at finding an appropriate representation to assess multilingual similarity. With this goal in mind, we look for a representation as target-independent as possible.
%
% why it is important
Similarity assessment is at the core of many natural language processing and information retrieval tasks.
Paraphrase identification is essentially similarity assessment and so is the task of plagiarism detection~\cite{Potthast:2010col}.
In multi-document summarisation~\cite{Goldstein:2000} finding two highly-similar pieces of information in two texts may imply it is worth adding them into a good summary. 
In information retrieval, particularly in question answering~\cite{hirschman_gaizauskas_2001}, a high similarity between a document and an information request is a key factor of relevance.
Similarity assessment also plays an important role in MT\@. It is essential in MT evaluation and, in the current cross-language setting, to identify parallel corpora to feed machine translation models~\cite{munteanuEtal:2005}.
Efforts have been carried out to approach cross-language versions of these tasks using interlingua or multilingual representations instead of translating the texts into one common language~\cite{Bouma:08,Munoz:08,Potthast:11}. Still, such representations are usually hard to design. This is where our neural context vector NMT embedding representation comes into play.
A multilingual encoder offers an environment where interlingua representations are learnt in a multilingual context. To some extent, it can be thought of as a generalisation of methods that project monolingual embeddings in two different languages into a common space to obtain bilingual word embeddings \cite{mikolovEtal:2013,faruquiDyer:2014,MadhyasthaEspana:2017}.

% related similar works
Recently, the authors in \cite{mccannEtAl:2107} used the context vectors (CoVe) from a deep LSTM encoder in a bilingual NMT system to complement GloVe word vectors\cite{penningtonEtAl:2014} and improve the performance on several tasks: sentiment analysis, question classification, entailment, and question answering. In their case, the purpose is to exploit the context of a word rather than the interlingual nature of its representation.
Finally, in a concurrent work, \cite{schwenkDouze:2017} describe how joint multilingual sentence representations are learned with an NMT architecture with multiple encoders and/or decoders. In their case, a sentence is represented by the last state of an LSTM or by the max pooling after a BLSTM, depending on the nature of the encoder. They go beyond a visual analysis and evaluate the equivalence among representations of the same sentence in different languages by looking at the error when recovering multilingual parallel corpora.

%% Data and System description
%% JSTSP Special Issue on End-to-End Speech and Language Processing

\section{NMT Systems Description}
\label{s:data}

\begin{table}[t]
 \caption{Description of the multilingual NMT systems. In all cases we use a learning rate of 0.0001, Adadelta optimisation, BPE vocabulary size of $2\,K$, 512-dimensional word embeddings,  mini-batch size of 80, and no drop-out.}
 \label{tab:NMTsistems}
 \centering
 \begin{tabular}{l ll cc}
  \toprule
       & Languages & Factor &  Hidden &  Vocabulary \\
       &  &  &  Units&   \\
  \midrule
      \sysarw & \{\ar, \en, \es\} &  word & $1024$ & $60\,K$ \\
      \sysarl & \{\ar, \en, \es\} & lemma & $1024$ & $60\,K$  \\
      \sysfrwdl & \{\de, \en, \es, \fr\} &  word &  $\,\,\,512$ & $80\,K$ \\
      \sysfrwdm & \{\de, \en, \es, \fr\} &  word & $1024$ & $80\,K$ \\
      \sysfrwdh & \{\de, \en, \es, \fr\} &  word & $2048$ & $80\,K$ \\
  \bottomrule
 \end{tabular}

\vspace{3mm} 
 \begin{subtable}{0.5\textwidth}   
\centering
\caption{Parallel sentences used in the \{\ar, \en, \es\} engine.}
\label{tab:nmt-corpora1}
\begin{tabular}{l  rrr}
\toprule
										& \ar--\en	& \ar--\es	& \en--\es	\\
\midrule
Training sentences	\\
\,\,\,\,\,United Nations~\cite{uncorpus}  					& $9.7\,M$	& $10.0\,M$	& $11.2\,M$	\\
\,\,\,\,\,Common Crawl%
\footnote{\url{http://commoncrawl.org}}						& --		& --		& $1.8\,M$	\\
\,\,\,\,\,News Commentary%
\footnote{\url{http://www.casmacat.eu/corpus/news-commentary.html}}		& $83\,K$	& $78\,K$	& $239\,K$	\\
\,\,\,\,\,IWSLT%
\footnote{\url{https://sites.google.com/site/iwsltevaluation2016/mt-track}}	& $90\,K$	& --		& --\\
\,\,\,\,\,Total									& $9.8\,M$	& $10.0\,M$	& $13\,M$	\\
\midrule
Validation Sentences	\\
\,\,\,\,\,newstest2012%
\footnote{\url{http://www.statmt.org/wmt14/translation-task.html}}		& --	& --	& $1.5\,K$		\\
\,\,\,\,\,eTIRR%
\footnote{LDC2004E72 available from the Linguistic Data Consortium}		& $1\,K$	&	--	& --	\\
\,\,\,\,\,News Commentary							& --	& $1\,K$	& --	 	\\
\bottomrule
\vspace{-5mm}
\end{tabular}   
\end{subtable}

\vspace{3mm}
\begin{subtable}{0.45\textwidth}
\caption{Parallel sentences used in the \{\de, \en, \es, \fr\} engine.}
\label{tab:nmt-corpora2}
\centering
\begin{tabular}{l @{\hspace{0mm}} rrrrrr}
\toprule
							& \de--\en	& \es--\en	& \fr--\en	& \es--\fr	\\
\midrule
Training sentences	\\
\,\,\,\,\, United Nations~\cite{multiun:2012}		& $162\,K$	& $11.2\, M$	& $12.9\, M$	& $11.6\, M$	\\
\,\,\,\,\, Common Crawl					& $2.4\, M$	& $1.8\, M$	& $3.2\, M$	& $1.1\, M$	\\
\,\,\,\,\, Europarl~\cite{europarl:2005}		& $1.9\, M$	& $2.0\, M$	& $2.0\, M$	& $1.9\, M$	\\
\,\,\,\,\, EMEA~\cite{tiedemann:2009}			& $1.1\, M$ 	& $1.1\, M$	& $1.1\, M$	& $394\, K$	\\
\,\,\,\,\, Scielo\footnote{\url{http://www.scielo.org}}	& --	 	& $676\, K$	& $9.0\, K$	& --	\\
\,\,\,\,\, Total					& $15M^*$	& $14M$		& $16M$		& $15\,M$	\\
\midrule

Validation Sentences	\\
\,\,\,\,\, newstest2012					& $22\,K$	& $22\,K$	& $22\,K$	& $22\,K$	\\
\bottomrule
  & \multicolumn{4}{l}{* Value obtained by oversampling}	\\
  \vspace{-7mm}
\end{tabular}
\end{subtable}

\end{table}

We carried out experiments with two multilingual many-to-many NMT engines trained with Nematus~\cite{nematus}. As in~\cite{johnsonEtal:2016} and similarly to \cite{haEtal:2016}, we trained our systems on parallel corpora for several language pairs L$i$--L$j$ simultaneously, adding a tag in the source sentence to account for the target language ``<2L$j$>'' (e.g., <2ar> if the target language is Arabic). Table~\ref{tab:NMTsistems} shows the key parameters of the engines. 
Since our aim is to study the capability of NMT representations to characterise similar sentences within and across languages, we selected languages for which text similarity and/or translation test sets are available. 

First, we build a ML-NMT engine for \ar, \en, and \es. We trained the multilingual system for the 6 language pair directions on $56\,M$ parallel sentences; see Table~\ref{tab:nmt-corpora1}.
We used 1024 hidden units, which correspond to 2048-dimensional context vectors.
We train system \sysarw\ after cleaning and tokenising the texts. 
A second system called \sysarl\ is trained on lemmatised sentences.
We used MADAMIRA~\cite{PashaEtAl:2014} for tokenisation and lemmatisation in \ar. For \en\ and \es\ we used Moses for tokenisation and IXA pipeline~\cite{AgerriEtAl:2014} for lemmatisation. 
In both cases we employ a vocabulary of $60\,K$ tokens plus $2\,K$ for subword units, segmented using Byte Pair Encoding (BPE)~\cite{sennrichEtala:2016}.

Second, we build a ML-NMT engine for \de, \fr, \en, and \es. We train the system with data on 4 language pairs: \de--\en, \fr--\en, \es--\en\ and \es--\fr. Although some corpora exist for the remaining two (\es--\de\ and \fr--\de), we exclude them to study these pairs as instances of zero-shot translation. 
We obtain $\sim$$15\,M$ parallel sentences per language pair ---for \de--\en, we oversampled to reach that amount by tripling the original sentences; see Table~\ref{tab:nmt-corpora2}.
We use a larger vocabulary in this engine: $80\,K$ type tokens plus $2\,K$ for BPE, as it involves one more language than in the first system. Only tokenisation with Moses is carried out.
Regarding the number of hidden units, we experiment with three configurations: \sysfrwdl, \sysfrwdm, and \sysfrwdh.
In all cases we used sentences no longer than 50 tokens.

For evaluation, we consider three types of test sets. The source side is always the same and is aligned to a target set that contains either: 
\Ni literal translations of the source, \Nii highly-similar sentences (both mono- and cross-language), and \Niii unrelated sentences (both mono- and cross-language).
For \ar, \en, and \es\ we build the three kinds of pairs out of the \textit{Semantic Textual Similarity Task at SemEval 2017} (STS 2017)~\cite{Agirre-EtAl-2017:SemEval}%
\footnote{\url{http://alt.qcri.org/semeval2017/task1}}. 
The task asks to assess the similarity between two texts within the range $[0,5]$, where $5$ stands for semantic equivalence.
We extract the subset of sentences with the highest similarity, 4 and 5, and use 140 sentences originally derived from the Microsoft Research Paraphrase Corpus~\cite{DolanEtAl:2005} (MSR), and 203 sentences from  WMT2008%
\footnote{\url{http://www.statmt.org/wmt08/shared-evaluation-task.html}} 
to build our final test set with 343 sentences (\sts).  These data were available for \ar\ and \en\ but not for \es, so we manually translated the MSR part of the corpus into \es, and gathered the \es\ counterparts of WMT2008 from the official set. With this process, we generated the test with translations ($\trad$) and highly similar sentence pairs ($\sims$). We shuffled one of the sides of the test set to generate the unrelated pairs ($\unrel$).

We use the test set from WMT2013 (\news) to simultaneously evaluate the \de, \fr, \en, and \es\ experiments; the last edition that includes these four languages. The test set contains $3 K$ sentences translated into the four languages. As before, we shuffle one of the sides to obtain the test set with unrelated sentence pairs, but we could not generate the equivalent set with highly similar pairs.
%% Analysis on the internal repesentation of an NMT multilingual system
%% JSTSP Special Issue on End-to-End Speech and Language Processing

\section{Context Vectors in Multilingual NMT Systems}
\label{s:vectors}

The NMT architecture used for the experiments is the encoder--decoder model with recurrent neural networks and attention mechanism described in Section~\ref{s:related}, as implemented in Nematus.
We use the sum of the context vector associated to %of 
every word (Eq.~\ref{e:ctx_word}) at a specific point of the training as the representation of a source sentence $s$:
\begin{equation}
  \mathbf{C} = \sum_{i=1}^n c_i.
  \label{e:ctx}
\end{equation}
This representation depends on the length of the sentence. However, we stick to this definition rather than using a mean over words
because the length of the sentences is a feature one might take into account, since sentences with similar meaning tend to have similar lengths.
Given sentence $s_1$ represented by $\mathbf{C}_{s_1}$ and sentence $s_2$ represented by $\mathbf{C}_{s_2}$, we can estimate their similarity by means of the cosine measure: 
\begin{equation}
  sim(\mathbf{C}_{s_1}, \mathbf{C}_{s_2}) = \frac{\mathbf{C}_{s_1} \cdot \mathbf{C}_{s_2}}{\Arrowvert\mathbf{C}_{s_1}\Arrowvert \, \Arrowvert\mathbf{C}_{s_2}\Arrowvert}.
  \label{e:cosine}
\end{equation}
By using this similarity measure we cancel the effect of the length of the sentence on the similarity between pairs but not on the representation of the sentence itself.\footnote{We explored alternative sentence representations (sum vs mean) and similarity measures (cosine vs modified versions of weighted Jaccard similarity, and Kullback--Leibler and Jensen--Shannon divergences). Cosine over the mean resulted in the best performance as measured by the correlation with human judgements on similarity assessments.}

\subsection{Graphical Analysis}
\label{ss:visual}

% Context vectors are high-dimensional structures: for the standard 1024-dimensional hidden layers one has 2048-dimensional context vectors.
Context vectors are high-dimensional structures: commonly used 1024-dimensional hidden layers lead to 2048-dimensional context vectors.
% , so they are difficult to visualise without a previous mapping into a low dimensional space. 
In order to get a first impression on the behaviour of the embeddings, we project the vectors for a set of sentences into a 2D space using t-Distributed Stochastic Neighbour Embedding (t-SNE)~\cite{VanDerMaaten:2014}.

\setcode{utf8}
\begin{figure}%[!t]
  \centering
    \fbox{\parbox{0.47\textwidth}{%
      \scriptsize
      \newcommand{\litem}[1]{\item{\hspace*{-0.5em}:$t$#1~~~~ }  }
      \itemsep=2pt       
      \begin{enumerate}[label=~~$s$\arabic*]
      \litem{1}  Spain princess testifies in historic fraud probe
      \litem{1}  Princesa de España testifica en juicio histórico de fraude
      \itemsep=0pt
      \litem{1}  \RL{‪أميرة‬ ‪أسبانيا‬ ‪تدلي‬ ‪بشهادتها‬ ‪في‬ ‪قضية‬ ‪احتيال‬ ‪تاريخي‬‪.‬}  
      \litem{2}  You do not need to worry.
      \itemsep=2pt
      \litem{3}  You don't have to worry. 
      \litem{2}  No necesitas preocuparte.
      \litem{3}  No te tienes por que preocupar.
      \litem{2}  \RL{‪لا‬ ‪ينبغي‬ ‪أن‬ ‪تقلق‬ }
      \itemsep=-2pt
      \litem{3}  \RL{‪لا‬ ‪ينبغي‬ ‪أن‬ ‪تجزع‬‪.‬}
       \itemsep=2pt
      \litem{4}  Mandela's condition has `improved'
      \litem{5}  Mandela's condition has `worsened over past 48 hours'
      \litem{4}  La salud de Mandela ha `mejorado'
      \litem{5}  La salud de Mandela `ha empeorado en las últimas 48 horas'
      \itemsep=0pt
      \litem{4}  \RL{‪لقد‬ ‪تحسنّت‬ ‪حالة‬ ‪مانديلا‬ ‪الصحية‬‪.‬ }
       \itemsep=-1pt
      \litem{5}  \RL{‪ساءت‬ ‪الحالة‬ ‪الصحية‬ ‪لمانديلا‬ ‪خلال‬ ‪ال‬ ‪48‬ ‪ساعة‬ ‪الماضية‬.‬ }
      \itemsep=2pt
      \litem{6}  Vector space representation results in the loss of the order \\ \hspace*{2.2em} which the terms are in the document.
      \litem{7}  If a term occurs in the document, the value will be non-zero \\ \hspace*{2.2em} in the vector.
      \litem{6}  La representación en el espacio de vecores implica la pérdida \\ \hspace*{2.2em} del órden en el que los  términos ocurren en el  documento.
      \litem{7}  Si un t\'ermino ocurre en el document, el valor en el vector \\ \hspace*{2.2em} será distinto de cero.
      \itemsep=0pt
      \litem{6}  \RL{‪يؤدي‬ ‪تمثيلُ‬ ‪فضاءِ‬ ‪المتجهِ‬ ‪إلى‬ ‪فقد‬ ‪الترتيب‬ ‪الذي‬ ‪تكون‬ ‪عليه‬ ‪المصطلحات‬ ‪في‬ ‪الوثيقة‬‪.‬}
      \itemsep=-2pt
      \litem{7}  \RL{‪إذا‬ ‪ما‬ ‪ورد‬ ‪مصطلح‬ ‪في‬ ‪الوثيقة‬‪,‬ ‪فالقيمة‬ ‪ستكون‬ ‪غيرصفرية‬ ‪المتجه‬‪.‬}
     \end{enumerate}     
   }}
   \caption{Set of 21 sentences chosen for the graphical analysis. The number of sentence $s$ and triplet $t$ used in subsequent plots is shown on the left-hand side. Sentences within a triplet have the exact same meaning (they are literal translations in \{\ar, \en, \es\}). Triplets ($t2$, $t3$), ($t4$, $t5$) and ($t6$, $t7$) share topic; hence they are close semantically.}
   \label{fig:trialSentences}
\end{figure}

\begin{figure*}[!t]
\centering
\includegraphics[width=0.32\textwidth]{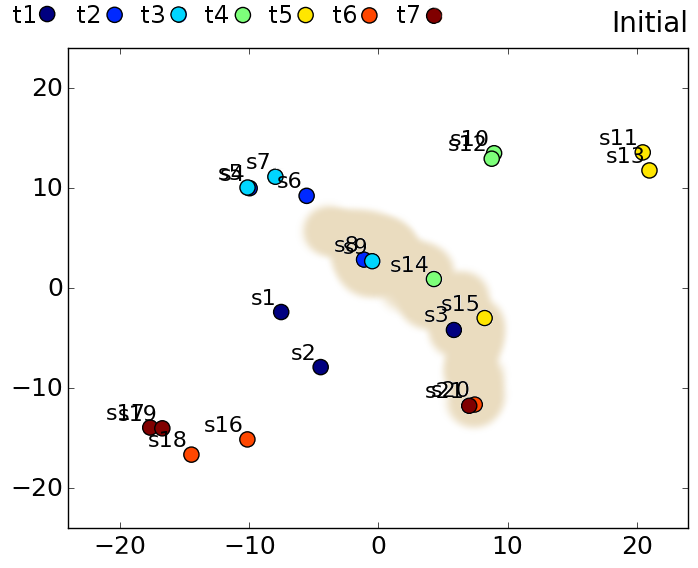}
\includegraphics[width=0.32\textwidth]{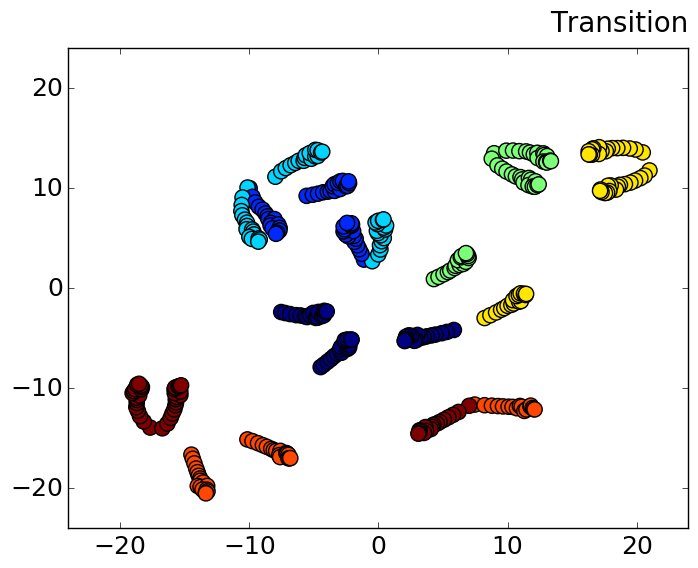}
\includegraphics[width=0.32\textwidth]{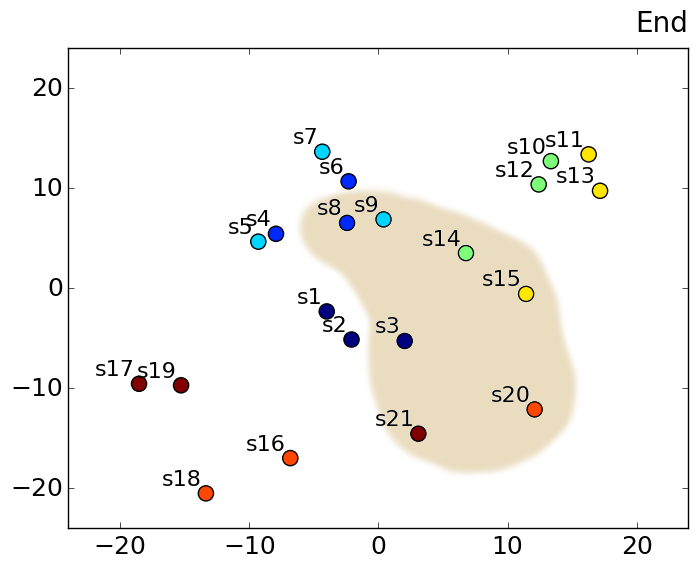}
\caption{2D t-SNE representation of the context vectors of the 21 sentences in Figure~\ref{fig:trialSentences}, obtained with the multilingual \{\ar, \en, \es\} NMT system, \sysarw. The left-most plot shows the vectors 
at a quite early stage of the training (after $10\cdot10^6$ sentences) and the right-most plot shows the vectors after 1.5 epochs ($178\cdot10^6$ sentences). The evolution during training is plotted in the middle panel. Shadowed regions include only Arabic sentences.}
\label{fig:vectors}
\end{figure*}

Figure~\ref{fig:trialSentences} shows 21 sentences extracted from the trial set of STS 2017 for this purpose and the relations between triplets.
Some triplets are related semantically; e.g., a triplet with the element ``\emph{Mandela's condition has improved}'' is semantically related to the triplet with the element ``\emph{Mandela's condition has worsened over past 48 hours}''. 
In a real multilingual space, one would expect sentences within a triplet to lie together and sentences within related triplets to be close but, as Figure~\ref{fig:vectors} shows, the range of behaviours may be diverse. The plot shows the evolution of the context vectors for these 21 sentences throughout the training (central panel), paying special attention to an early (left panel) and a late stage (right panel). 

At the beginning of the training, \en\ and \es\ sentences in the same triplet (same colour) lie close together and even overlap for some triplets; e.g.,~$t4$ and $t7$. This is an effect of having a representation that depends on the length of the sentence: the elements in $t4$ and $t7$ not only share some vocabulary, but also have very similar lengths. Arabic sentences remain together, almost irrespective of their meaning. 
One has to take into account that \en\ and \es\ are closer between them than to \ar.
Meanwhile, \ar\ is closer to \es\ than to \en.
% At this early training stage, the closer languages have already been clustered together (\en\ and \es)
At this early training stage, the closer languages already cluster together (\en\ and \es)
and sentences can be grouped according to their semantics, but the most distant language (\ar) is not in the same stage yet. 
At this stage, pairs where both sentences are written in \ar\ are considered more similar, even if they are semantically very different (also compared to semantically similar sentences across languages); sentence $s9$ is closer to $s14$ (another sentence in \ar\ with similar length) than to $s7$ (a strict and longer translation of $s9$ into \es). 

As training continues, \ar\ sentences spread through the space and slowly tend to join their counterparts in the other languages. English and Spanish sentences also move apart towards a more general interlingua position. That is, there is a flow from near to overlapping locations for translations of the same sentence towards locations grouped by topic, irrespective of the language (e.g.,~see the evolution of the related triplets $t6$ and $t7$). 
This evolution must be considered if one wants to use context vectors as a semantic representation of a sentence: representations at different points of the training process might be useful for different tasks. 
For instance, as shown in the following subsections, using context vectors from a converged NMT training is beneficial to assess similarity, but one only needs to run some iterations to have appropriate vectors to identify parallel sentences.

However, not all the triplets show the expected behaviour. 
While  at every iteration the sentences in the triples in $t1$ and $t5$ each move closer together, and therefore behave as expected, the sentences in $t6$ move further away from each other (notice that this triplet has the longest sentences and the highest length variation). 
A more systematic study is necessary in order to be able to draw strong conclusions. In the following sections we conduct such a study and draw conclusions quantitatively, rather than only qualitatively. 

\subsection{Source vs Source--Target Semantic Representations}
\label{ss:sourceML}

The training of the ML-NMT systems involves one-to-many instances. That is, for the same source language L1 one has different examples of translations into L2, L3, or L4. A first question one can address given this setup is whether the interpretation of a source sentence learnt by the network depends on the language it is going to be translated into or not. In a truly interlingual space, such representations should be the same, or at least very close.
To test this, we compute the cosine similarity between the representation of a source sentence $s$ when it is translated with the same engine into two different languages L$i$ and L$j$:
\begin{equation}
 {\rm<2L}i-{\rm2L}j> \equiv sim(s_{{\rm<2L}i>}, s_{{\rm<2L}j>}) \enspace.
\end{equation}
Sentence representations are extracted with engine \sysarw\ for \{\ar, \en, \es\} on \sts\ data and with engine \sysfrwdm\ for  \{\de, \en, \es, \fr\} on \news. Afterwards, we compute the mean over all the sentences in a test set.

\begin{table}%[t]
 \caption{Similarities between the internal representations of the sentences in \sts\ (sys. \sysarw) and \news\ (sys. \sysfrwdm) when translated from L1 into different languages {L2, L3, L4}.  1$\sigma$ uncertainties are shown in parentheses and affect the last significant digit; similarities appear starred when a zero-shot language pair is involved.}
 \label{tab:source}
 \centering
   \begin{tabular}{cc ccc}
     \toprule
       L1 & \{L2, L3, L4\} & <2L2--2L3> 	& <2L2--2L4> & <2L3--2L4>\\
     \midrule
      \ar & \{\en,\es,$\phi$\} & \,\, $0.97(5)$ & -- & -- \\
      \en & \{\es,\ar,$\phi$\} & \,\, $0.94(5)$ & -- & -- \\
      \es & \{\ar,\en,$\phi$\} & \,\, $0.91(5)$ & -- & -- \\
     \midrule
      \de & \{\fr,\en,\es\} & *$0.97(2)$ 	& *$0.98(2)$ 	& *$0.96(2)$ \\
      \fr & \{\en,\es,\de\} & \,\, $0.96(2)$	& *$0.96(2)$ 	& *$0.97(2)$ \\
      \en & \{\es,\de,\fr\} & \,\, $0.96(2)$	& \,\, $0.98(2)$& \,\, $0.96(2)$ \\
      \es & \{\de,\fr,\es\} & *$0.97(2)$	& *$0.96(2)$ 	& \,\, $0.97(2)$ \\
     \bottomrule
   \end{tabular}

\end{table}

\begin{table*}[t]
 \caption{Cosine similarities between the obtained representations of the sentences in the \sts\ test set with \sysarw\ and \sysarl. The results are shown for both monolingual and cross-language language pairs and the three sets with translations ($\trad$), semantically similar sentences ($\sims$) and unrelated sentences ($\unrel$). Notice that a $\trad$ set cannot be built in the monolingual case. 
 \deltaTrUr\ is the difference between the mean similarity seen in translations and in unrelated sentences. 1$\sigma$ uncertainties are shown in parentheses and affect the last significant digits.}
 \label{tab:basicARENES}
 \centering

 \begin{tabular}{@{\hskip 0.1em}l@{\hskip 0.4em}l@{\hskip 1.5em}l ccccc @{\hskip 3em} ccccc}
  \toprule
         & &   & \multicolumn{5}{c}{\sysarw ords} & \multicolumn{5}{c}{\sysarl emmas}\\
              \cmidrule(lr){4-8}  \cmidrule(lr){9-13}
         & &   & \ar--\ar        & \en--\en     & \ar--\en        & \ar--\es        & \en--\es
               & \ar--\ar        & \en--\en     & \ar--\en        & \ar--\es        & \en--\es \\
   \midrule
    \tiny  \multirow{5}{*}{\rot{\scriptsize 0.1 EPOCHS~}} & 
           \multirow{5}{*}{\rot{\scriptsize ($4\cdot10^6$sent.)~}} & 
    \\
    & & $\trad$ &  -- & -- & $0.26(10)$ & $0.76(05)$ & $0.40(09)$ 
            &  -- & -- & $0.44(07)$ & $0.81(04)$ & $0.53(05)$ \\
   
    & & $\sims$ & $0.92(03)$ & $0.93(01)$ & $0.24(10)$ & $0.75(06)$ & $0.38(09)$ 
            & $0.93(01)$ & $0.94(01)$ & $0.42(07)$ & $0.80(05)$ & $0.51(06)$ \\
    
    & & $\unrel$& $0.65(13)$ & $0.66(13)$ & $0.06(09)$ & $0.53(11)$ & $0.14(10)$ 
            & $0.70(09)$ & $0.73(09)$ & $0.27(09)$ & $0.63(10)$ & $0.33(08)$ \\
            
     \cmidrule(lr){3-13}
    & & \deltaTrUr &   -- & --  & $0.20(13)$ & $0.23(12)$ & $0.26(13)$ 
               &   -- & --  & $0.16(11)$ & $0.18(11)$ & $0.20(10)$ \\
             
  \midrule
    \tiny  \multirow{5}{*}{\rot{\scriptsize 0.5 EPOCHS~}} & 
           \multirow{5}{*}{\rot{\scriptsize ($28\cdot10^6$sent.)}} & 
    \\
    & & $\trad$ &  --        &       --   & $0.61(07)$ & $0.67(06)$ & $0.76(06)$
            &  --        &       --   & $0.51(06)$ & $0.68(05)$ & $0.60(06)$\\
    
    & & $\sims$ & $0.86(07)$ & $0.87(06)$ & $0.58(08)$ & $0.65(07)$ & $0.73(07)$
            & $0.84(08)$ & $0.86(06)$ & $0.47(07)$ & $0.66(07)$ & $0.57(07)$\\
    
    & & $\unrel$ & $0.48(12)$ & $0.43(12)$ & $0.30(10)$ & $0.37(11)$ & $0.37(11)$
             & $0.45(12)$ & $0.46(11)$ & $0.23(08)$ & $0.39(10)$ & $0.27(09)$\\

    \cmidrule(lr){3-13}
    & & \deltaTrUr &   -- & --  & $0.32(12)$ & $0.30(12)$ & $0.39(12)$ 
               &   -- & --  & $0.28(11)$ & $0.29(11)$ & $0.33(11)$ \\

   \midrule
   \tiny  \multirow{5}{*}{\rot{\scriptsize 1.0 EPOCHS~}} & 
           \multirow{5}{*}{\rot{\scriptsize ($56\cdot10^6$sent.)}} & 
    \\
    & &   $\trad$ &  --        &       --   & $0.61(08)$ & $0.65(07)$ & $0.74(06)$
            &  --        &       --   & $0.51(06)$ & $0.63(06)$ & $0.60(06)$ \\

    & & $\sims$ & $0.83(09)$ & $0.85(07)$ & $0.57(08)$ & $0.63(08)$ & $0.70(08)$
            & $0.81(10)$ & $0.83(07)$ & $0.47(07)$ & $0.61(08)$ & $0.56(07)$ \\

    & & $\unrel$ & $0.41(12)$ & $0.37(11)$ & $0.27(10)$ & $0.32(11)$ & $0.31(10)$
             & $0.38(12)$ & $0.40(11)$ & $0.21(08)$ & $0.33(09)$ & $0.25(09)$ \\
    \cmidrule(lr){3-13}
    & & \deltaTrUr &   -- & --  & $0.34(12)$ & $0.33(13)$ & $0.43(12)$ 
               &   -- & --  & $0.28(11)$ & $0.29(11)$ & $0.33(11)$ \\

  \midrule
   \tiny  \multirow{6}{*}{\rot{\scriptsize 2.0 EPOCHS~}} & 
          \multirow{6}{*}{\rot{\scriptsize ($112\cdot10^6$sent.)}} & 
    \\
    & & $\trad$  &  --        &       --   & $0.59(07)$ & $0.62(07)$ & $0.71(07)$
             &  --        &       --   & $0.50(06)$ & $0.60(06)$ & $0.59(07)$\\

    & & $\sims$ & $0.80(10)$ & $0.83(08)$ & $0.54(08)$ & $0.60(08)$ & $0.67(08)$ 
            & $0.78(11)$ & $0.82(08)$ & $0.46(07)$ & $0.58(08)$ & $0.56(08)$ \\
    
    & & $\unrel$ & $0.37(12)$ & $0.34(11)$ & $0.26(09)$ & $0.30(10)$ & $0.29(10)$
             & $0.33(11)$ & $0.36(10)$ & $0.21(08)$ & $0.29(08)$ & $0.22(08)$\\
   \cmidrule(lr){3-13}
    & & \deltaTrUr &   -- & --  & $0.33(12)$ & $0.32(12)$ & $0.42(12)$ 
              &   -- & --  & $0.29(10)$ & $0.31(10)$ & $0.37(11)$ \\
   \vspace{-0.7em} \\
 \bottomrule
 \end{tabular}
\end{table*}

Table~\ref{tab:source} shows the results. 
The similarities are close to $1$ in all cases, a number that would indicate that the representations are fully equivalent, and are compatible with $1$ within a $2\sigma$ interval.
Although the differences among languages and test sets are not significant at that level, some general trends are observed. Despite the fact that the similarity between instances of the same sentence is not $1$, it is larger than the similarity between closely related sentences when translated into the same language (see Section~\ref{ss:mainAnalysis}); i.e.\ we can identify a sentence by a unique representation.
Also notice that there is no difference when we translate into a language without any direct parallel data (zero-shot translation): system \sysfrwdm\ had no data for \es--\de\ and \fr--\de, but 
the similarities involving these pairs (starred in Table~\ref{tab:source}) are not statistically-significantly different from those involving \es--\fr\ and \es--\en, for example.

Finally, we can strengthen the correlation of the relatedness between languages and the closeness of the internal representations observed also via the first graphical analysis. 
The representation of an \ar\ sentence when translated into \en\ or \es\ is almost the same ($sim=0.97\pm0.05$), but the difference in the representation of an \es\ sentence when translated into \ar\ or \en\ is the largest one ($sim=0.91\pm0.05$) due to the disparity between \ar\ and \en. The same effect is observed in \{\de, \fr, \en, \es\} at a lower degree when making the distinction between \{\fr, \es\} and \{\de, \en\} as two groups of  ``close'' languages.

\subsection{Representations throughout Training}
\label{ss:mainAnalysis}

During training, the network learns the most appropriate representation of words/sentences in order to be translated, so the embeddings themselves evolve over time. As seen in the graphical analysis (Section~\ref{ss:visual}), it is interesting to follow this evolution and examine how sentences are grouped together depending on their language and semantics. Hence, we analyse in parallel an engine trained on lemmatised sentences (\sysarl) and one trained on tokenised sentences (\sysarw). 
The rationale is that the vocabulary in the lemmatised system is smaller and therefore can be better covered by the $60\,K$ NMT fixed vocabulary during training. Still, the ambiguity becomes higher, which could damage the quality of the representations.

Table~\ref{tab:basicARENES} shows the results. At the beginning of the training process, after having seen $4\cdot10^6$ sentences only, the results are still very much dependent on the language. Translations in \ar--\es\ have a similarity of $0.81\pm0.04$, whereas translations in \ar--\en\  have a similarity of $0.44\pm0.07$ (first row for system \sysarl emmas). Perhaps for this reason monolingual pairs show higher similarity values than cross-language pairs, even for unrelated sentences ($sim = 0.70\pm0.09$ for \ar\ and $sim = 0.73\pm0.09$ for \en). 
Nevertheless, within a language pair the system is already aware of the meaning of the sentences: cosine similarities are the highest for translations ($\trad$), slightly lower for semantically related sentences ($\sims$) and significantly lower for unrelated sentences ($\unrel$). The difference between the mean similarities obtained for translations and unrelated sentences,
$$ \Delta_{\rm tr-ur} \equiv \Delta(sim(\trad) - sim(\unrel)),$$
shows that, already at this point, parallel sentences can be identified and located in the multilingual space, even though the similarity for translations is in general far from $1$ and the similarity for unrelated sentences is far from $0$.
In the worst-case scenario, S1-lemmas for \ar--\en, $\Delta_{\rm tr-ur}=0.16\pm0.11$, so translations are clearly distinguished at $1\sigma$ level. In other words, if we look at the distance of one sentence to its translation and to all the unrelated sentences in the $unrel$ set, only in $1.6\%$ of the cases an unrelated sentence is closer or at the same distance as the translation. This number  diminishes to $0.6\%$ in the best case scenario (S1-words for \en--\es).
Also at this starting point, sentences lie closer together irrespective of their meaning in the lemmatised system than in the tokenised one. 
Similarities are always higher for \sysarl\ than for its counterpart in \sysarw. The separation between translations and unrelated sentences is always more important in the \sysarw\ (\deltaTrUr\ is higher). 
This is true all along the training process, supporting the hypothesis that the ambiguity introduced by the lemmatisation damages the representativeness of the embeddings.

\begin{table}[t]
 \caption{Akin to Table~\ref{tab:basicARENES} for the \{\de, \fr, \en, \es\} engine on the \news\ test sets after half an epoch. In this case, three system configurations are shown that vary the size of the last hidden layer of the encoder: \sysfrwdl, \sysfrwdm\ and \sysfrwdh.}
 \label{tab:basicDEENESFR}
 \centering

 \begin{tabular}{@{\hspace{0em}}l@{\hspace{0.5em}}l@{\hspace{0.5em}}
                c@{\hspace{0.7em}}c@{\hspace{0.7em}}c@{\hspace{0.7em}}c@{\hspace{0.7em}}c@{\hspace{0.7em}}c@{\hspace{0em}}}
  \toprule
           &         & \de--\en        & \de--\es            & \de--\fr        & \en--\es        & \en--\fr    & \es--\fr \\
  \midrule
   \multicolumn{6}{@{\hspace{0.1em}}l}{\ssysfrwdl}\\
           & $\trad$  & $0.61(10)$ & $0.62(10)$ & $0.62(10)$ & $0.66(10)$ & $0.66(10)$ & $0.73(10)$ \\
           & $\unrel$ & $0.25(10)$ & $0.27(10)$ & $0.27(10)$ & $0.26(10)$ & $0.26(10)$ & $0.30(11)$ \\
             \cmidrule{2-8}
  & \deltaTrUr & $0.36(14)$ & $0.35(14)$ & $0.35(14)$ & $0.40(14)$ & $0.41(14)$ & $0.43(15)$ \\
 \midrule
   \multicolumn{6}{@{\hspace{0.1em}}l}{\ssysfrwdm}\\
           & $\trad$  & $0.62(10)$ & $0.62(10)$ & $0.62(10)$ & $0.66(10)$ & $0.66(10)$ & $0.73(10)$ \\
           & $\unrel$ & $0.26(10)$ & $0.27(10)$ & $0.27(10)$ & $0.26(10)$ & $0.27(10)$ & $0.31(11)$ \\
             \cmidrule{2-8}
 & \deltaTrUr & $0.36(14)$ & $0.35(14)$ & $0.34(14)$ & $0.39(14)$ & $0.40(14)$ & $0.42(15)$ \\
  \midrule
   \multicolumn{6}{@{\hspace{0.1em}}l}{\ssysfrwdh}\\
           & $\trad$  & $0.59(10)$ & $0.58(10)$ & $0.58(10)$ & $0.61(10)$ & $0.62(10)$ & $0.69(11)$ \\
           & $\unrel$ & $0.24(09)$ & $0.25(09)$ & $0.25(09)$ & $0.23(09)$ & $0.23(09)$ & $0.27(10)$ \\
             \cmidrule{2-8}
 & \deltaTrUr & $0.35(13)$ & $0.33(14)$ & $0.33(14)$ & $0.38(13)$ & $0.39(14)$ & $0.42(15)$ \\
  \bottomrule
 \end{tabular}

\end{table}

When the training process has covered $28\cdot10^6$ sentences, half an epoch for this system, the difference among languages diminishes. Now sentences lie closer together in the tokenised system than in the lemmatised one, irrespective of their meaning.
From this point onwards, this trait is maintained. Although all similarities keep going down throughout the training, even for translations, \deltaTrUr\ remains almost constant. The maximum value for this difference is found after one epoch ($\sim56\cdot10^6$ sentences) for all the cross-language pairs in the tokenised system. In this case, \deltaTrUr\ is $0.34\pm0.12$ for \ar--\en, $0.33\pm0.13$ for \ar--\es\ and $0.43\pm0.12$ for \en--\es. Again, the distinction is the clearest for the closest language pair and diminishes when \ar\ is involved, mainly because translations involving \ar\ are more difficult to detect (the mean similarity between \en--\es\ translations is $0.74\pm0.06$; $0.61\pm0.08$ for \ar--\en).

As Table~\ref{tab:basicDEENESFR} shows, analogous conclusions can be drawn from the \{\de, \fr, \en, \es\} engine. The maximum distinction between related and unrelated sentences \deltaTrUr\ is found after $\sim56\cdot10^6$ sentences, half an epoch in this case, even though the difference was well established at one third of an epoch.
\deltaTrUr\ is $0.3\pm0.1$ when \de\ is involved (\de--\en, \de--\es, \de--\fr) and $0.4\pm0.1$ when not (\en--\es, \en--\fr, \es--\fr). The difference is mostly given by the similarity between translations, which is higher when \de\ is not concerned. 

\begin{figure}[!t]
\centering
\includegraphics[width=0.8\columnwidth]{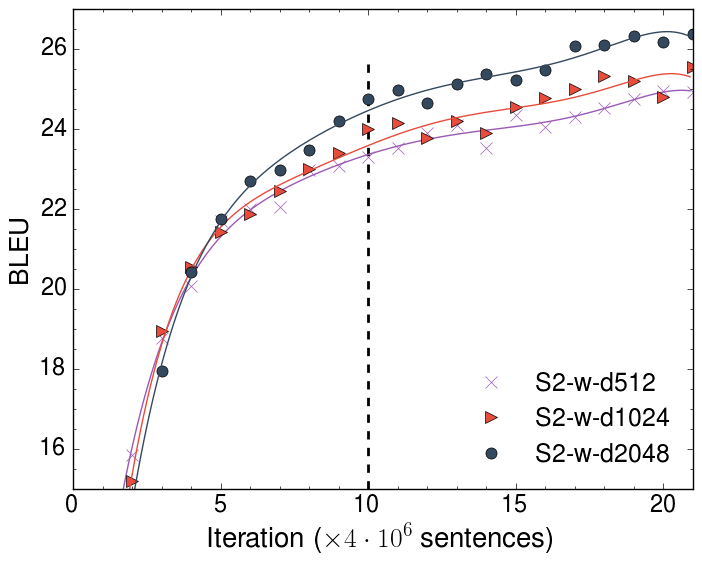}
\caption{BLEU evolution throughout training on \news\ when translated from English into Spanish with three systems that differ in the size of the hidden layer (see text). The vertical line marks the point where context vectors achieve the maximum descriptive power.}
\label{fig:bleus}
\end{figure}

Notice that this optimal point does not correspond to the optimal point regarding translation quality. Figure~\ref{fig:bleus} displays the progression of the BLEU score along training for the \en2\es\ translation. The dashed vertical line indicates the iteration where \deltaTrUr\ is maximum. At this time, the engine is still learning, as reflected by the fact that the translation quality is clearly increasing.
Another interesting observation is that the expressiveness of the embeddings does not depend on their dimensionality. Context vectors with 1024 dimensions (\sysfrwdl), 2048 dimensions (\sysfrwdm) and 4096 dimensions (\sysfrwdh), lead to similar figures for similarity values between pairs of sentences. At the beginning of the training, \sysfrwdm\ gives slightly better representations than the other two systems, but this difference is narrowed when the training evolves. The training time almost doubles when doubling the dimensionality of the hidden layer, but this higher capacity does not result in a better description of the data. Indeed, 4096-dimensional vectors perform worse than the 1024-dimensional ones at all the training stages. However, translation quality does depend on the size of the hidden layer and, in our experiments, \sysfrwdh\ performs better than the lower-dimensional systems. 

\subsection{Similarity Assessments}
\label{ss:semeval}

Up to now, we have mostly analysed how similar ($trad$) and dissimilar ($unrel$) sentences behave across languages and during training. The degree of similarity was left aside because the $trad$ and $semrel$ test sets are too alike to draw statistically-significant conclusions in that setting. To do so, we evaluate the use of context vectors as a feature to assess similarities in the STS framework.
In this case, we use all the available test sets for the 2017 evaluation campaign with sentence pairs ranging from completely unrelated sentences (score 0) to semantic equivalents (score 5). Only the subset of most similar sentences had been used in the earlier experiment (scores 4 and 5).

Table~\ref{tab:semeval} shows the Pearson correlation between the predictions given by the context vectors of \sysarw\ and \sysarl\ and human assessments for five language pairs. Observing the evolution through training by taking a shot at four different points, the correlation increases with the number of iterations for all the language pairs and systems. In this fine-grained task, the internal representation improves in parallel to the translation quality. As before, the system with words is better than the one with lemmas with the only exception of \ar--\en. A reason could be the low initial similarity for semantically equivalent sentences ($trad$) for this pair with the \sysarw\ system ($0.26\pm0.10$).
The initial point seems to be relevant for the final performance; i.e.\ the relative improvement from epoch to epoch for all language pairs is very similar, but the final performance seems to be conditioned to the quality of the initial representations. The performance in the monolingual tracks is always higher than in the cross-language ones, and the difference at the end of the training is proportional to the difference at the beginning.
The study of how a proper initialisation of the input word embeddings could alleviate this disparity deserves future research.

The comparison with word vector embeddings obtained with the word2vec skip-gram model~\cite{mikolovEtal:2013:ICLR} is specially interesting. We estimated 300 (\weCCCnmt) and 1024 (\weMXXIVnmt) dimensional word embeddings with the same corpus used to train the NMT systems (adding monolingual corpora did not improve the results). When sentences belong to different languages, we translate them into \en\ and use the embeddings estimated for \en. As done with context vectors, the similarity between sentences is assessed by the cosine of the summed embeddings.  Higher-dimensional word embeddings outperform the 300-dimensional ones in the task. Yet, even with the 1024-dimensional word embeddings, the performance is far from that obtained with context vectors ---between $0.04$ and $0.21$ points lower (see last block of Table~\ref{tab:semeval}).

\begin{table}[t]
 \caption{Comparison of the Pearson correlation obtained by context vectors at different epochs of the training and word embeddings on the test set of the \textquotedblleft Semantic Textual Similarity Task'' at SemEval 2017.}
 \label{tab:semeval}
 \centering
  \begin{tabular}{lcccccc}  
  \toprule
                & track1 & track2 & track3 & track4a & track5\\
                & \ar--\ar & \ar--\en & \es--\es & \es--\en & \en--\en\\
  \midrule
  \sysarw-0.1Ep & 0.32 & 0.25 & 0.55 & 0.32 & 0.54 \\
  \sysarw-0.5Ep & 0.52 & 0.36 & 0.71 & 0.40 & 0.68 \\
  \sysarw-1.0Ep & 0.57 & 0.42 & 0.74 & 0.44 & 0.72 \\
  \sysarw-2.0Ep & 0.59 & 0.44 & 0.78 & 0.49 & 0.76 \\
  \midrule                                           
  \sysarl-0.1Ep & 0.29 & 0.32 & 0.50 & 0.25 & 0.49 \\
  \sysarl-0.5Ep & 0.49 & 0.45 & 0.67 & 0.38 & 0.65 \\
  \sysarl-1.0Ep & 0.53 & 0.51 & 0.71 & 0.42 & 0.69 \\
  \sysarl-2.0Ep & 0.57 & 0.54 & 0.75 & 0.45 & 0.73 \\
  \midrule                                           
  \weCCCnmt     & 0.49 & 0.28 & 0.55 & 0.40 & 0.56\\
  \weMXXIVnmt   & 0.51 & 0.33 & 0.59 & 0.45 & 0.60\\
  \bottomrule
  \end{tabular} 
\end{table}
%% Introduction for the article
%% JSTSP Special Issue on End-to-End Speech and Language Processing

\section{Use Case: Parallel Sentence Extraction}
\label{s:adaptation}

The previous section showed how ML-NMT context vectors can be used as a representation to calculate sensitive similarities between sentences with the potential to distinguish translations from non-translations and even translations from pairs with similar meaning. Among other applications, we can use the representations learned when mapping parallel sentences ---the NMT system training--- to detect new parallel pairs. 
Now we use a semantic similarity measure based on the context vectors obtained with the NMT system of Section~\ref{s:vectors} to extract parallel sentences and study its performance compared to other measures. 
Our translation engine is the ML-NMT \{\de, \fr, \en, \es\} system described in Section~\ref{s:data}. After the conclusions gathered in Section~\ref{s:vectors}, we use system \sysfrwdl\ after half an epoch of training to extract the context vectors. This system gives the best trade-off between speed (low-dimensional vectors are extracted faster) and dissociation between translations and unrelated sentences, as this is the training point where the difference \deltaTrUr\ is maximum. 

In order to perform a complete analysis, we consider five complementary measures to context vectors and test different scenarios. 
We borrow two well-known representations from cross-language information retrieval to account for syntactic features by means of cosine similarities:
\Ni character $n$-grams~\cite{mcnamee2004character} with $n=[2,5]$ and  \Nii pseudo-cognates. 
From a natural language point of view, cognates are ``words that are similar across languages''~\cite{Manning-Schutze:1999}. We relax the concept and consider as pseudo-cognates any words in two languages that share prefixes. To do so, tokens shorter than four characters are discarded, unless they contain non-alphabetical characters. The resulting tokens are cut down to four characters~\cite{simard1993using}.
The preprocessing consists only of casefolding and punctuation/diacritics removal. For the character $n$-gram measure, we also remove spaces to better account for compounds in German.
We also include general features at sentence level such as \Niii token and \Niv character counts, and \Nv the length factor measure~\cite{journals/corr/abs-cs-0609060}. 

We test three different scenarios to observe the effect of context vectors when extracting sentence pairs and compare them against the other standard characterisations: \vspace{-1mm}
\begin{description}
 \item[$ctx$:] only context vectors,
 \item[$comp$:] only the set of five complementary measures, and
 \item[$all$:] a combination of $ctx$ and $comp$.
\end{description}
\vspace{-1mm}

\noindent
For each scenario, we learn a binary classifier on annotated data. We use the \de--\en\ and \fr--\en\ training corpora provided for the shared task on identifying parallel sentences in comparable corpora at BUCC 2017~\cite{Zweigenbaum:17}.%
\footnote{\url{https://comparable.limsi.fr/bucc2017/bucc2017-task.html}}
This set contains $1.5\,M$ sentences from Wikipedia and News Commentary from which $20\,K$ are aligned sentence pairs. 
Negative indexes are manually added by randomly pairing up the same amount of non-matching pairs to build a balanced data set. We use $35\,K$ instances from the full set for training and evaluating classifiers with 10-fold cross-validation, $4\,K$ instances for training an ensemble of the best classifiers and $1\,K$ instances for held-out testing purposes.

For $ctx$, where only the context vector similarities are considered, the problem can be reduced to finding a suitable decision threshold. To this end, similarity values between the lowest value among positive examples and the highest value among negative samples are incrementally increased by a step size of 0.005 and the threshold giving the highest accuracy on the training set is selected. 
With this methodology, we obtain a threshold $t=0.43$ for \de--\en\ leading to an accuracy of $97.2\%$, and $0.41$ for \fr--\en\ with an accuracy of $97.4\%$. These values are slightly lower than the ones reported in Table~\ref{tab:basicDEENESFR}, but consistent with them. The thresholds in both cases depend on the language pair, but the fact that we are working with an interlingua representation makes the differences minimal. 
In such a case, one can estimate a joint threshold for the full training set in \de--\en\ and \fr--\en\, and later use this decision boundary for other language pairs.
If we do the search on the joint datasets the best threshold is $t=0.43$ leading to an accuracy of $97.2\%$ in the training set.

We have 7 and 8 features in $comp$ and $all$ and employ supervised classifiers rather than a threshold estimation: 
support vector machines (SVM) with RBF kernel and gradient boosting (GB) on the deviance objective function with 10-fold cross-validation. A soft voting ensemble (Ens.) of SVM and GB is trained to obtain the final model.\footnote{We use the \textit{Python} \textit{scikit-learn} package: \url{http://scikit-learn.org}}

\begin{table}
\caption{Precision, recall and F$_1$ scores on the binary classification of pseudo-alignments on the held-out test set.}
\label{tab:extraction}
\centering
\begin{tabular}{@{\hspace{0em}}l@{\hspace{0.7em}}l c@{\hspace{0.7em}}c@{\hspace{0.7em}}c c@{\hspace{0.7em}}c@{\hspace{0.7em}}c c@{\hspace{0.7em}}c@{\hspace{0.7em}}c@{\hspace{0em}}}
  \toprule
     &  & \multicolumn{3}{c}{\de--\en} & \multicolumn{3}{c}{\fr--\en} & \multicolumn{3}{c}{joint}\\
             \cmidrule(lr){3-5}  \cmidrule(lr){6-8} \cmidrule(lr){9-11}
     &  & P & R & F$_1$ & P & R & F$_1$ & P & R & F$_1$\\
  \midrule
     \parbox[t]{2mm}{\multirow{4}{*}{\rotatebox[origin=c]{90}{$ctx$}}} 
      & Thrs. & 95.5 &\bf 97.1  &96.3	& 95.4& \bf 100.0&  97.7	& 98.3	& 98.1	& 98.2\\
      & SVM   & 96.2 &	96.2    &96.2	& 95.6&  99.1	 &  97.3	& 97.1  & 98.0	& 97.6\\
      & GB    & 97.0 &	95.7    &96.4	& 95.6&  99.6  	 &  97.6	& 97.0  & 97.3	& 97.2\\
      & Ens.  & 98.2 &	95.7    &97.0	& 95.6&  99.1	 &  97.3	& 96.9  & 97.8	& 97.3\\

  \midrule
    \parbox[t]{2mm}{\multirow{3}{*}{\rotatebox[origin=c]{90}{$comp$}}}      
      & SVM  &72.3&	85.5&	78.4&    76.7&	85.1&	80.7&    73.4&	80.9&	77.0 \\
      & GB   &93.5&	85.1&	89.1&    97.2&	93.2&	95.1&    96.9&	90.7&	93.7 \\
      & Ens. &84.0&	89.4&	86.6&    95.5&	95.5&	95.5&    93.4&	91.6&	92.5 \\
  \midrule
    \parbox[t]{2mm}{\multirow{3}{*}{\rotatebox[origin=c]{90}{$all$}}}     
      & SVM  & 74.6&	86.4&	80.1	&    81.8   &	87.3&	84.5   &   86.1   &	85.6&	85.8\\
      & GB   & 98.7&	96.6&	\bf 97.6&  \bf 99.1 &	99.6& \bf 99.3 & \bf 98.9 &	98.9&	\bf 98.9 \\
      & Ens. & \bf 99.1 & 96.6&	\bf 97.8 & \bf 99.1 &	99.6& \bf 99.3 &    98.7  & \bf 99.1 &	\bf 98.9 \\
  \bottomrule
\end{tabular}
\end{table}

Table~\ref{tab:extraction} shows precision (P), recall (R) and F$_1$ scores for the three scenarios.
Notice that a greedy threshold search is better than any of the machine learning counterparts when only context vectors are used, but differences are not significant. 
The greedy search on the context vector similarities gives a better F$_1$ on the held-out test set than an ensemble of SVM and GB operating only the set of additional features with almost no knowledge of semantics.
As we argued in the previous section, translations and non-translations are clearly differentiated by a cosine similarity of the context vectors for these languages pairs, as the difference between the mean similarities of translations and unrelated texts is much higher than its  uncertainty (\deltaTrUr $=0.36\pm0.14$ for \de--\en, and $0.41\pm0.14$ for \fr--\en). This clear distinction in the similarities is translated into an F$_1=98.2\%$ in the task of parallel sentence identification.

Due to its interlingual nature, our feature behaves equally well for both language pairs and improves in the multilingual setting (Table~\ref{tab:extraction}, joint columns). By contrast, the set of complementary features depends on the language pair and shows a performance drop for \de--\en. For this reason, the results in the multilingual setting are always worse than in the bilingual one. This fact is inherited in the $all$ scenario, where the classification for the joint corpus obtains F$_1=98.9\%$, which is lower than the one obtained for \fr--\en\ alone (F$_1=99.3\%$). 
Nevertheless, semantic and syntactic similarity features are complementary and the combination of all similarity measures slightly improves precision, recall and F$_1$ in the multilingual setting.
It is worth noting the high recall derived from the context vectors, which reaches $100\%$ for  \fr--\en\, and falls to $98.1\%$ for the joint data, being still 6.5 points higher than for the $comp$ features.  
%% Conclusions for 
%% JSTSP Special Issue on End-to-End Speech and Language Processing

\section{Conclusions}
\label{s:conclusions}

In this article we provide evidence of the interlingual nature of the context vectors generated by a multilingual neural machine translation system and study their power in the assessment of mono- and cross-language similarity. 
Comparisons with word vectors show that context vectors are able to capture better the semantics in the two settings. 

The study addresses four main research questions, introduced in Section~\ref{s:intro}. Regarding RQ1, we investigate how the representation of a sentence varies in order to be accommodated to a particular target language and observe that the difference is negligible, even though it grows when we consider distant target languages, such as Arabic and English. Even in these cases, the representation of a sentence is unique enough as closely related sentences have a lower similarity than different instances of the same sentence.
RQ2: The results also show that the context vectors are able to differentiate among sentences with identical, similar, and different meaning  across different languages ---Arabic, English, French, German, and Spanish. The difference between translations and non-translations can be established at least at $1\sigma$ level for all the pairs. As a direct application, we identify parallel sentences in comparable corpora, obtaining $F_1=98.2\%$ on data of the shared task at BUCC 2017. 
The correlation of the cosine between context vectors with human judgements on continuous similarity assessments ranges in $[0.4,0.8]$, always higher than the ones obtained for word vectors models: $[0.3,0.6]$.
RQ3: The language dependence is not completely lost in the representations. In the latter experiment, correlations in the cross-language tasks are lower than in the monolingual ones, but in both cases related and unrelated sentence pairs are clearly distinguishable within the variance. RQ4: Our training-evolution experiments reveal that the first feature to locate a sentence in the multilingual space is its language but, after only $\sim$$4\cdot10^6$ training sentences, the model is already aware of the semantics. As the training evolves, the difference between translations and unrelated sentences grows till reaching a plateau when the system has been trained on $\sim$$40\cdot10^6$ sentences.
Vectors at early training are therefore already adequate for identifying parallel sentences, whereas the optimal ones for fine-grained similarity assessments and translation require further training. 

Given these conclusions, several research avenues are worth exploring in the future.
The disparity in the performance of mono- and cross-language similarity assessment tasks
triggers a question on how relevant the initialisation of the embeddings is. Could the results be improved with initialisations of the word embeddings other than random? The answer can be extended and exploited in other natural language processing tasks, in the same 
% triggers various questions. How relevant the initialisation of the embeddings is? Could the results be improved with initialisations of the word embeddings other than random? Additionally, the study can be extended to other natural language processing tasks, in the same 
philosophy as~\cite{mccannEtAl:2107}, but in a multilingual setting.
Additionally, similar studies using other NMT architectures could help in better understanding the insights of the learning.

% or

% \appendices
% % you can choose not to have a title for an appendix
% % if you want by leaving the argument blank
% \section{}

\ifCLASSOPTIONcompsoc
  \section*{Acknowledgments}
\else
  \section*{Acknowledgment}
\fi

% \begin{footnotesize}
This work was partially funded by the Leibniz Gemeinschaft via the SAW-2016-ZPID-2 project
and by the European Union Horizon 2020 research and innovation programme under grant agreement 645452 (QT21).
The research of A. Barr\'on-Cede\~no is carried out in the framework of the Interactive  sYstems  for  Answer  Search  project (IYAS) at QCRI\@.
We thank the anonymous reviewers for their interesting insights that helped to improve this paper.
% \end{footnotesize}

\ifCLASSOPTIONcaptionsoff
  \newpage
\fi


% \cris{References: TOCHECK}

\bibliographystyle{IEEEtran}
\bibliography{nmt}

\end{document}